\documentclass{article} 
\usepackage[table]{xcolor}
\usepackage{iclr2024_conference,times}

\usepackage{amsmath,amsfonts,bm}

\def\eqref#1{equation~\ref{#1}}

\def\1{\bm{1}}

\DeclareMathAlphabet{\mathsfit}{\encodingdefault}{\sfdefault}{m}{sl}
\SetMathAlphabet{\mathsfit}{bold}{\encodingdefault}{\sfdefault}{bx}{n}

\DeclareMathOperator*{\argmin}{arg\,min}

\usepackage{hyperref}
\usepackage{url}

\title{Optimizing for ROC Curves on Class-Imbalanced Data by Training over a Family of Loss Functions}

\author{Kelsey Lieberman, Shuai Yuan, Swarna Kamlam Ravindran \& Carlo Tomasi \\
Department of Computer Science\\
Duke University\\
Durham, NC 27708 USA \\
}

\usepackage{graphicx}
\usepackage{subcaption}
\usepackage{multirow}
\usepackage[capitalize,noabbrev]{cleveref}
\usepackage{float}
\usepackage{wrapfig}
\newcommand{\ie}{{\em i.e.}}
\newcommand{\eg}{{\em e.g.}}

\iclrfinalcopy 
\begin{document}

\maketitle

\begin{abstract}
Although binary classification is a well-studied problem in computer vision, training reliable classifiers under severe class imbalance remains a challenging problem. Recent work has proposed techniques that mitigate the effects of training under imbalance by modifying the loss functions or optimization methods. While this work has led to significant improvements in the overall accuracy in the multi-class case, we observe that slight changes in hyperparameter values of these methods can result in highly variable performance in terms of Receiver Operating Characteristic (ROC) curves on binary problems with severe imbalance. To reduce the sensitivity to hyperparameter choices and train more general models, we propose training over a family of loss functions, instead of a single loss function. We develop a method for applying Loss Conditional Training (LCT) to an imbalanced classification problem. Extensive experiment results, on both CIFAR and Kaggle competition datasets, show that our method improves model performance and is more robust to hyperparameter choices. Code is available at \url{https://github.com/klieberman/roc_lct}.



\end{abstract}

\section{Introduction}
\label{sec:intro}


Consider a classifier which takes images of skin lesions and predicts whether the lesions are melanoma \cite{rotemberg2020patientcentric}. Such a system could be especially valuable in underdeveloped countries where expert resources for diagnosis are scarce \cite{Cassidy2022-rc}. Classifying melanoma from images is a problem with class imbalance since benign lesions are far more common than melanomas. Furthermore, the accuracy on the melanoma (minority) class is much more important than the accuracy on the benign (majority) class because predicting a benign lesion as melanoma would result in the cost of a biopsy while predicting a melanoma lesion as benign could result in the melanoma spreading before the patient can receive appropriate treatment. 

In this case, overall accuracy, even on a balanced test set, is clearly an inadequate metric, as it implies that the accuracies on both classes are equally important. Instead, Receiver Operating Characteristic (ROC) curves are better suited for such problems \cite{roc_book}. These curves plot the tradeoff between the true positive rate (TPR) on the y-axis and the false positive rate (FPR) on the x-axis over a range of classification thresholds. Unlike scalar metrics (\eg, overall accuracy on a balanced test set or $F_\beta$), ROC curves show model performance over a wide range of classification thresholds. This allows practitioners to understand how the model's performance changes based on different classification thresholds and choose the best tradeoff for their needs. ROC curves can also be summarized by their Area Under the Curve (AUC) \cite{roc_book}. Furthermore, both ROC curves and AUC have several mathematical properties which make them preferred to alternative precision-recall curves \cite{Flach2015-wg}\footnote{We provide definitions and visualizations of commonly-used metrics for binary problems with imbalanced data in Appendix~\ref{sec:appendix_metrics}}. 

Although binary problems, like melanoma classification, are often cited as the motivation for class imbalance problems and ROC curves are the \textit{de facto} metric of choice for such problems, the class imbalance literature largely focuses on improving performance on longtailed multi-class datasets in terms of overall accuracy on a balanced test set. We instead focus on binary problems with severe imbalance and propose a method, which adapts existing techniques for handling class imbalance, to optimize for ROC curves in these binary scenarios.


\begin{figure}
    \centering
    \includegraphics[width=0.5\linewidth]{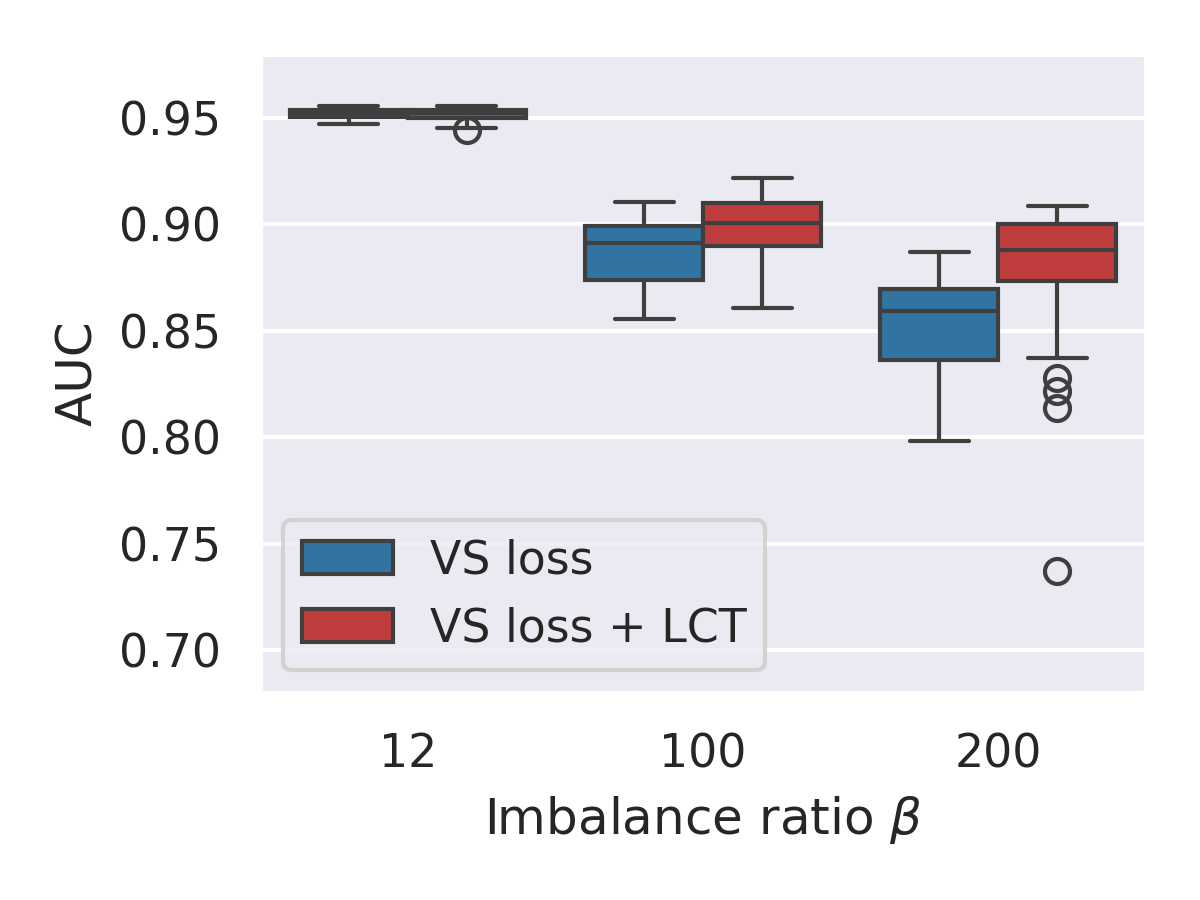}
    \caption{Distribution of Area Under the ROC Curve (AUC) values obtained by training the same model on the SIIM-ISIC Melanoma classification dataset with 48 different combinations of hyperparameters on VS Loss (hyperparameter values are given in Section~\ref{sec: experiments}). Results are shown at three different imbalance ratios. \textbf{As the imbalance becomes more severe, model performance drops and the variance in performance drastically increases. LCT addresses both of these issues by training over a family of loss functions, instead of a single loss function with one combination of hyperparameter values.}}
    \label{fig:melanoma_boxplots}
\end{figure}


In particular, we adapt Vector Scaling (VS) loss, which is a general loss function for imbalanced learning with strong theoretical backing \cite{kini2021labelimbalanced}. VS loss is a modification of Cross-entropy loss that adjusts the logits via additive and multiplicative factors. There is theory supporting the use of both of these factors: multiplicative factors are essential for the terminal phase of training, but these have negative effects early during training, so additive factors are necessary speed up convergence. Although VS loss has shown strong performance in the multi-class setting, it does have hyperparameters which require tuning (\eg, the additive and multiplicative factors on the loss function).



We find that, in the binary case, the effect of these hyperparameters is small and reasonable at moderate imbalance ratios $\beta$ (where $\beta$ is defined as the ratio of majority to minority samples); however, at large imbalance ratios, small differences in these hyperparameters lead to very wide variance in the results (Figure~\ref{fig:melanoma_boxplots}). This figure shows that increasing the imbalance ratio not only decreases the AUCs (as expected), but also drastically increases the variance in AUCs obtained by training models with slightly different hyperparameter values. 

In this work, we highlight the practical effect of the theoretically-motivated VS loss on the ROC metric, especially for data problems with high imbalance ratios. We propose a method that adapts VS loss to align the training objective more closely with ROC curve optimization. Our method trains a single model on a wide range of hyperparameter values using Loss Conditional Training \cite{Dosovitskiy2020You}. We find that this method not only reduces the variance in model performance caused by hyperparameter choices, but also improves performance over the best hyperparameter choices since it optimizes for many tradeoffs on the ROC curve.  We provide extensive results-- both on CIFAR datasets and datasets of real applications derived from Kaggle competitions-- at multiple imbalance ratios and across a wide range of hyperparameter choices.


In summary, our contributions are as follows.
\begin{itemize}
    \item We identify that higher levels of imbalance are not only associated with worse model performance, but also more variance.
    \item We recognize that training over a range of hyperparameter values can actually benefit classification models that are otherwise prone to overfitting to a single loss function. 
    \item We propose using Loss Conditional Training (LCT) to improve the training regimen for classification models trained under imbalance.
    \item We show that this method consistently improves performance at high imbalance ratios.
\end{itemize}


\section{Related work}
\label{sec: related_work}


Many solutions have been proposed to address class imbalance, including several specialized loss functions and optimization methods \cite{cao2019learning, rangwani2022escaping, Buda2018-tl, kini2021labelimbalanced, shwartz-ziv2023simplifying}. Perhaps the simplest of these is to change the class weights in the loss function so that the minority and majority class have ``balanced'' class weights or class weights which are inversely proportional to the frequency of the class in the training set (\eg, weighted cross-entropy loss \cite{Xie1989-iy}). Another popular loss function is Focal loss, which down-weights ``easy'' samples (\ie, samples with high predictive confidence) \cite{Lin2017-gk}.


More recently, several loss functions have been proposed which add additive and multiplicative factors to the logits before they are input to the softmax function \cite{cao2019learning, Ye2020-yj, menon2021longtail}. These aim to enforce larger margins on the minority class and/or calibrate the models. \citet{kini2021labelimbalanced} recognized that many of the previous loss functions for addressing class imbalance can be expressed by one general loss function: Vector Scaling (VS) loss, which gives strong performance on multi-class datasets after hyperparameter tuning \cite{kini2021labelimbalanced}. \citet{du2023global} use a global and local mixture consistency loss, contrastive learning, and a dual head architecture. 


Additionally, \citet{rangwani2022escaping} proposed using Sharpness Aware Minimization (SAM) as an alternative optimizer and found that this helped the model escape saddle points in multi-class problems with imbalance \cite{foret2021sharpnessaware}. Similarly, \citet{shwartz-ziv2023simplifying} identify several tweaks that can be made to models---including batch size, data augmentation, specialized optimizers, and label smoothing---which can all improve training.

These methods are predominantly tested on multi-class datasets with overall accuracy on a balanced test set as the primary metric. We instead propose to optimize for the ROC curve in the binary case by training over a family of loss functions. To this end, we use Loss Conditional Training (LCT), which was proposed as a way to train one model to work over several rates in a variable-rate problem \cite{Dosovitskiy2020You}. LCT was proposed for applications such as neural image compression and variational autoencoders and, to our knowledge, we are the first to use it to improve ROC curves of a classification problem.

\section{Problem setup}

\subsection{Data}
In this paper, we focus on binary classification problems with class imbalance.
Specifically, let $D = \{(\mathbf{x}_i, y_i)\}_{i=1}^n$ be the training set consisting of $n$ \textit{i.i.d.} samples from a distribution on $\mathcal{D}=\mathcal{X}\times \mathcal{Y}$ where $\mathcal{X} = \mathbb{R}^d$ and $\mathcal{Y} = \{0, 1\}$.
Then let $n_c$ be the number of samples with label $c$.
Without loss of generality, we assume that $n_1 < n_0$ (\ie, 0 is the majority class and 1 is the  minority class) and measure the amount of imbalance in the dataset by $\beta=n_0/n_1>1$.

\subsection{Predictor and predictions}
Let $f$ be a predictor with weights $\boldsymbol{\theta}$ and $\mathbf{z}=f(\mathbf{x}, \boldsymbol{\theta})$ be $f$'s output for input $\mathbf{x}$. We assume $\mathbf{z}$ are logits (\ie, a vector $(z_0, z_1)$ of unnormalized scalars where $z_c$ is the logit corresponding to class $c$) and $\mathbf{p}$ are the outputs of the softmax function (\ie, a vector $(p_0, p_1)$ of normalized scalars such that $p_0 + p_1 = 1$). Then the model's prediction is 
\begin{align}
    \hat{y} = \begin{cases}
        1 &\text{if } p_1 > t, \\
        0 &\text{otherwise},
    \end{cases}
\end{align}
where $t \in [0, 1]$ is a threshold and $t=0.5$ by default. To find the ROC curve of a classifier $f$, we compute the predictions over a range of $t$ values.

\subsection{Loss function}
\label{sec:setup_loss}

For all experiments, we use the Vector Scaling (VS) loss as defined by \citet{kini2021labelimbalanced}. This loss is a modification of weighted Cross-entropy loss with two hyperparameters $\Delta, \iota$ that specify an affine transformation $\Delta z + \iota$ for each logit:

\begin{align}\label{eq:vs_loss}
    \ell_{VS}(y, \mathbf{z}) = -\omega_y \log\left(\frac{e^{\Delta_y z_y + \iota_y}}{\sum_{c\in \mathcal{Y}} e^{\Delta_c z_c + \iota_c}}\right).
\end{align}

We follow parameterization by \citet{kini2021labelimbalanced} as follows:
\begin{align}\label{eq:tau_gamma}
    \iota_c = \tau \log \left(\frac{n_c}{n}\right) \quad\text{and}\quad
    \Delta_c = \left(\frac{n_c}{n_0} \right)^\gamma
\end{align}
where $\tau \geq 0$ and $\gamma \geq 0$ are hyperparameters set by the user. We add an additional hyperparameter $\Omega$ to parameterize each class' weight $\omega_c$ as follows
\begin{align}
    \omega_c = \begin{cases}
                \Omega & \text{if } c = 1, \\
                1-\Omega & \text{otherwise},
                \end{cases}
\end{align}
where $\Omega \in [0, 1]$. In the binary case, we can simplify the loss as follows (see Appendix~\ref{sec:appendix_simplifying_vs} for details):
\begin{align}
    \ell_{VS}(0, \mathbf{z}) &= (1-\Omega)\log\left(1+ e^{z_1/\beta^\gamma - (z_0 + \tau \log\beta)} \right)\label{eq:vs0} \\
    \ell_{VS}(1, \mathbf{z}) &= \Omega\log\left(1 + e^{(z_0 + \tau \log\beta) - z_1/\beta^\gamma)}\right).\label{eq:vs1}
\end{align}

Note that VS loss is equivalent to cross entropy loss when $\Omega=0.5, \gamma=0$, and $\tau=0$. It is also equal to weighted cross entropy loss when $\Omega=\frac{n_0}{n}, \gamma=0$, and $\tau=0$.
\section{Analysis on the effects of hyperparameters}
\label{sec:hyper}

\begin{figure}
    \centering
    \includegraphics[width=0.5\linewidth]{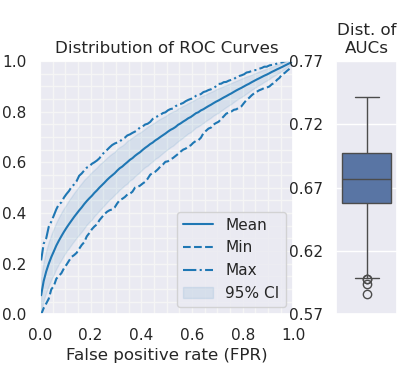}
    \caption{Distribution of results obtained from training 512 models with different hyperparameter values. \underline{Left:} mean, min, max, and 95\% confidence interval of ROC curves. \underline{Right:} boxplot of distribution of Area Under the ROC Curve (AUC) values.} 
    \label{fig:baseline_dist}
\end{figure}

\begin{figure*}
\centering
\hbox{\hskip0cm%
  \vtop{\vskip0pt%
    \hbox{%
         \includegraphics[width=0.85\linewidth]{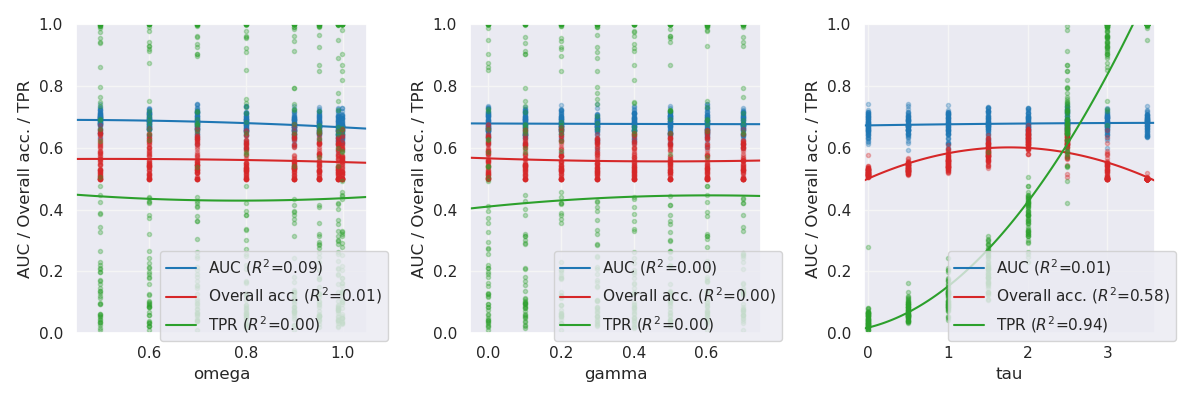}   
      }%
  }\hskip-62mm%
  \vtop{\vskip5mm
        \begin{tabular}{|c|c|}
        \hline
        Metric & $R^2$ \\
        \hline
            AUC & 0.11  \\
            Acc. & 0.60 \\
            TPR & 0.95 \\
            \hline
        \end{tabular}
  }%
}
\caption{Effect of VS loss hyperparameters on AUC, overall accuracy, and TPR. Results are shown for 512 models with different hyperparameter values. For each metric and hyperparameter we plot a) all the values of the metric vs. the hyperparameter (dots) and b) a fitted degree-2 polynomial between the metric and hyperparameter (curves). In the table, we report the $R^2$ values of polynomials fit with all three hyperparameters. All models were trained on CIFAR10 cat vs. dog with $\beta=100$ with a ResNet32 model. \textbf{Most of the variance in AUC cannot be explained by the hyperparameter values.}}
\label{fig:metrics_vs_hyper}
\end{figure*}

\subsection{Variance of results}
\label{sec:hyper_variance}


As discussed in Section~\ref{sec:setup_loss}, our parameterization of VS loss has three hyperparameters: $\Omega, \gamma$ and $\tau$. Each of these has theory associated with its effect on training under imbalance. $\Omega$ controls the overall balance between the magnitude of the gradients from the minority and majority class. $\gamma$ was originally proposed to compensate for the difference between the magnitude of the minority-class logits at training and testing time \cite{Ye2020-yj}; however, \citet{kini2021labelimbalanced} showed that it is essential for training beyond zero error and optimizing with $\gamma>0$ is equivalent to a solving Cost-Sensitive Support Vector Machine problem. $\tau$ enforces a larger margin on minority-class samples and has many of its own theoretical properties \cite{cao2019learning}: in particular, it has been shown to counteract some of the negative effects of training with $\gamma>0$ early in training \cite{kini2021labelimbalanced}. Although there are intuitive and theoretical rationale for each of these hyperparameters \cite{kini2021labelimbalanced}, it is not clear how sensitive a model's performance is to their values, especially in terms of ROC curves on a binary dataset.


To understand this better, we trained several models on the same dataset over different combination of hyperparameter values. Specifically, we trained 512 models on the same dataset (CIFAR10 cat vs. dog with $\beta=100$) and varied only their values of $\Omega, \gamma$ and $\tau$.
\footnote{We used all unique combinations of $\Omega \in \{$0.5, 0.6, 0.7, 0.8, 0.9, 0.95, 0.99, 0.999$\}$, $\gamma\in\{$0.0, 0.1, 0.2, 0.3, 0.4, 0.5, 0.6, 0.7$\}$, $\tau \in \{$0, 0.5, 1, 1.5, 2, 2.5, 3, 3.5$\}$.} \cref{fig:baseline_dist} shows that there is a wide variance in performance of these models
Specifically, these models have AUC values which range from 0.58 to 0.74 with a standard deviation of 0.03. Additionally, Figure~\ref{fig:melanoma_boxplots} shows that this variance becomes more severe as the imbalance ratio increases. \textbf{This implies that models trained with the VS loss  under severe class imbalance are especially sensitive and variable in their results.}




\subsection{Correlation between hyperparameter values and scalar metrics}
\label{sec:hyper_metric_correlation}


We next consider how much of this variance is explained by the hyperparameter values (\ie, is the randomness coming from the hyperparameters themselves or other randomness in training?). If the variance is mostly a result of different hyperparameter choices, then we can reasonably expect the practitioner to search for hyperparameters that generalize well. Otherwise, tuning hyperparameters may not be an effective way to improve performance.

To test this, we fit polynomials between the three hyperparameters ($\Omega, \gamma, \tau$) and two scalar metrics: AUC and overall accuracy on a balanced test set when $t=0.5$. Although TPR, equivalently recall, is not a good indicator of a model's overall performance by itself (because it is trivial to increase TPR at the expense of FPR), we also fit a polynomial to TPR to gain insights on the effect of the hyperparameters on an ROC curve. We show results, including the $R^2$ of these regressions, in the three plots in Figure~\ref{fig:metrics_vs_hyper}. We also fit three (one for AUC, accuracy, and TPR) degree-2 polynomials using all three hyperparameters as features and report $R^2$ values of these polynomials in the table in Figure~\ref{fig:metrics_vs_hyper}. We choose polynomials of degree two instead of linear regression because we expect that an optimum exists. 

We see that no single hyperparameter is strongly correlated with AUC; however, 11\% of the variance can be explained by the polynomial fit to all three hyperparameters. This suggests that some variance can be removed by choosing strong hyperparameters; however, much of the variance comes from other sources. Additionally, only 60\% of the variance in overall accuracy can be explained by the polynomial with all three hyperparameters. While this is significantly better than AUC's $R^2$, it is still far from perfect. 
\textbf{Thus, VS Loss hyperparameter values do not affect performance metrics in a very predictable way. It would be preferable to reduce the variance of VS Loss over performing extensive hyperparameter tuning.}


\subsection{Connection between hyperparameter values and softmax outputs}
\label{sec:hyper_tau}


The third plot in Figure~\ref{fig:metrics_vs_hyper} shows that $\tau$ is strongly correlated with TPR when the threshold $t=0.5$ ($R^2=0.95$). To understand this, recall that $\tau$ affects the additive term on the logits in \cref{eq:vs_loss,eq:tau_gamma}. Specifically, increasing $\tau$ enforces a larger margin on the minority class. In general, this will lead to increasing the softmax scores of the minority class at inference time, which is similar to post-hoc VS calibration \cite{zhao2020role}. Thus, when $t$ is constant, increasing $\tau$ will likely cause an increase in TPR.

\begin{figure}[tb]
    \centering
    \includegraphics[width=0.5\linewidth]{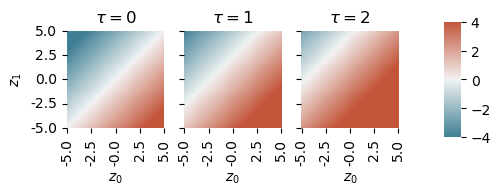}
    \caption{Effect of $\tau$ on loss landscape. Each plot shows $\ell_{VS}(1,\mathbf{z})- \ell_{VS}(0,\mathbf{z})$ over $z_0, z_1 \in [-5, 5]$ for $\beta=10$. White: ``Break-even'' points. \textbf{$\boldsymbol{\tau}$ shifts the loss landscape.}}
    \label{fig:break-even-tau}
\end{figure}

We can also visualize the effect of $\tau$ by considering the set of ``break-even'' logits, which are defined as the set of points of $\mathbf{z}$ where the loss is equal whether the sample has label 0 or 1 (\ie, the points of $\mathbf{z}$ such that $\ell_{VS}(1,\mathbf{z}) = \ell_{VS}(0,\mathbf{z})$). With regular Cross-entropy, this set is $z_1=z_0$. Assume instead that we keep $\Omega=0.5, \gamma=0$, but have arbitrary values for $\tau$, then the equation for the set of break-even points becomes $z_1 = z_0 + \tau \log\beta$. In other words, varying $\tau$ is equivalent to shifting the set of break-even points (and the rest of the loss landscape) by $\tau \log\beta$.\footnote{Varying $\tau$ also shifts the loss landscape in a similar way for any fixed $\Omega\in(0,1)$ and $\gamma>0$. See \cref{sec:appendix_break-even} for details.}. Figure~\ref{fig:break-even-tau} visualizes this.

In addition, $\Omega$ and $\gamma$ also have well-understood effects on the logit outputs. \cref{sec:appendix_break-even,sec:appendix_hyper_contour} include a general equation for the set of break-even points and more loss contours. Just as we saw in the example with $\tau$, different combinations of hyperparameters have different effects on how $\mathbf{z}$ is scaled, which in turn affects their softmax scores. \textbf{Thus, training with different values of VS Loss hyperparameters corresponds to optimizing for different tradeoffs of TPR and FPR.}

\section{Optimizing for ROC Curves via Loss Conditional Training}
\label{sec:method}

Recall that ROC curves show the performance of the model in terms of TPR for all possible FPRs and \textit{vice versa.} 
We saw in Section~\ref{sec:hyper_tau} that training with different combinations of hyperparameters corresponds to optimizing for different tradeoffs of TPR and FPR. Therefore, training one model over a range of hyperparameter values is a proxy for optimizing over a range of the ROC curve. Using this intuition, we design a system which optimizes one model over a range of VS loss hyperparameter values via loss conditional training (LCT). 




\subsection{Loss Conditional Training (LCT)}
\citet{Dosovitskiy2020You} observed the computational redundancy involved in training separate models with slightly different loss functions, such as neural image compression models with different compression rates. They proposed loss conditional training (LCT) as a method to train one model to work over a family of losses. 

Let $\boldsymbol{\lambda}$ be a vector which parameterizes a loss function. For example, we can parameterize VS loss by  $\boldsymbol{\lambda} = (\Omega, \gamma, \tau)$. Then let  $\mathcal{L}(\cdot, \cdot, \boldsymbol{\lambda})$ be the family of loss functions parameterized by $\boldsymbol{\lambda} \in \boldsymbol{\Lambda} \subseteq \mathbb{R}^{d_{\boldsymbol{\lambda}}}$ where $d_{\boldsymbol{\lambda}}$ is the length of $\boldsymbol{\lambda}$. In our example, $\mathcal{L}(\cdot, \cdot, \boldsymbol{\lambda})$ is the set of all possible VS loss functions obtained by different parameter values for $\Omega, \gamma, \tau$. 

Normally, training finds the weights $\boldsymbol{\theta}$ on a model $f$ which minimize a single loss function $\mathcal{L}(\cdot, \cdot, \boldsymbol{\lambda})$ (\ie, we optimize for a single combination of $\boldsymbol{\lambda}_0 = (\Omega_0, \gamma_0, \tau_0)$ values in VS loss) as shown below.
\begin{align}
\label{eq:one_loss}
    \boldsymbol{\theta}^*_{\lambda_0} &= \argmin_{\boldsymbol{\theta}} \mathbb{E}_{(\mathbf{x},y) \sim D} \mathcal{L}(y, f(\mathbf{x}, \boldsymbol{\theta}), \boldsymbol{\lambda}_0)
\end{align}

LCT instead optimizes over a distribution of $\boldsymbol{\lambda}$ values $P_{\boldsymbol{\Lambda}}$ as shown below.
\begin{align}
\label{eq:lct_loss}
    \boldsymbol{\theta}^* &= \argmin_{\boldsymbol{\theta}} \mathbb{E}_{\boldsymbol{\lambda} \sim P_{\boldsymbol{\Lambda}}} \mathbb{E}_{(\mathbf{x},y) \sim D} \mathcal{L}(y, f(\mathbf{x}, \boldsymbol{\theta}, \boldsymbol{\lambda}), \boldsymbol{\lambda})
\end{align}
LCT is implemented on Deep Neural Network (DNN) predictors by augmenting the network to take an additional input vector $\boldsymbol{\lambda}$ along with each data sample $\mathbf{x}$. During training, $\boldsymbol{\lambda}$ is sampled from $\boldsymbol{\Lambda}$ with each sample and is used in two ways: 1) as an additional input to the network and 2) in the loss function for the sample. During inference, the model takes as input a $\boldsymbol{\lambda}$ along with the sample and outputs the appropriate prediction.

In order to condition the model on $\boldsymbol{\lambda}$, the DNN is augmented with Feature-wise Linear Modulation (FiLM) layers \cite{Perez_2018_film}. These are small neural networks that take the conditioning parameter $\boldsymbol{\lambda}$ as input and output a $\boldsymbol{\mu}$ and $\boldsymbol{\sigma}$, used to modulate the activations channel-wise based on the value of $\boldsymbol{\lambda}$.
Specifically, suppose a layer in the DNN has an activation map of size $H \times W \times C$. 
In LCT, we transform each activation $\mathbf{f}\in\mathbb{R}^C$ by $\boldsymbol{\mu}$ and $\boldsymbol{\sigma}$ as follows: $\mathbf{\Tilde{f}} = \boldsymbol{\sigma} *\mathbf{f} + \boldsymbol{\mu}$ where both $\boldsymbol{\mu}$ and $\boldsymbol{\sigma}$ are vectors of size $C$, and ``$*$'' stands for element-wise multiplication. 


\subsection{Details about implementing LCT for classification}

We define our family of loss functions $\boldsymbol{\lambda}$ using VS loss. We try several choices for $\boldsymbol{\lambda}$, including $\boldsymbol{\lambda} = \Omega, \boldsymbol{\lambda} = \gamma, \boldsymbol{\lambda} = \tau$ and $\boldsymbol{\lambda} = (\Omega, \gamma, \tau)$. In the first three cases, we set the hyperparameters excluded from $\boldsymbol{\lambda}$ to constants (as is done in regular VS loss). We find that $\boldsymbol{\lambda} = \tau$ is especially strong at improving the average performance and reducing the variance in performance (Figure~\ref{fig:choice_of_lambda}), which is supported by our insights in \cref{fig:metrics_vs_hyper} and \cref{sec:hyper_tau}. Thus, for most of our experiments, we use $\boldsymbol{\lambda}=\tau$. 

For each mini-batch, we draw one $\boldsymbol{\lambda}$. This is done by independently sampling each hyperparameter in $\boldsymbol{\lambda}$ from a distribution that has a linear density over range $[a, b]$. In this distribution, the user specifies $a, b$, and the height of the probability density function (pdf) at $b$, $h_b$. The height at $a$, $h_a$, is then found to ensure the area under the pdf is 1. Unlike the triangular distribution, this general linear distribution does not require the pdf to be 0 at either endpoint and, unlike the uniform function, it does not require the slope of the pdf to be 0.\footnote{Appendix~\ref{sec:appendix_linear} contains more details about the linear distribution.} We use this general distribution as a way to experiment with different distributions.

Although the value of $\boldsymbol{\lambda}$ at inference time affects the TPR and FPR when $t=0.5$, we observe that the ROC curves are almost identical for all values of $\boldsymbol{\lambda}$ in the range the model was trained on (see Appendix~\ref{sec:eval_lambda}). Thus, we evaluate LCT models at one $\boldsymbol{\lambda}$ and find their ROC curve from this output.

\section{Experiments}
\label{sec: experiments}

\subsection{Experimental setup}

\underline{Methods.} For each dataset and imbalance ratio $\beta$, we train 48 models with the regular \textbf{VS loss}, varying their hyperparameter values. We also train 48 models with LCT applied to VS loss where $\boldsymbol{\lambda}=\tau$ (\textbf{VS Loss + LCT}). We use $\Omega=0.5, 0.7, 0.9, 0.99$, $\gamma=0.0, 0.2, 0.4$ for both methods (note that we set $\Omega, \gamma$ to constants when $\boldsymbol{\lambda}=\tau$ in LCT). We use $\tau=0, 1, 2, 3$ for VS loss, and $[a,b]=[0,3]$ with $h_b=0.0, 0.15, 0.33, 0.66$ for VS loss + LCT. We evaluate with  $\boldsymbol{\lambda}=\tau=3$. To apply LCT, we augment the networks with one FiLM block after the final convolutional layer and before the linear layer. This block is comprised of two linear layers with 128 hidden units. Specifically, the first layer takes one input $\boldsymbol{\lambda}$ and outputs 128 hidden values and the second layer outputs 64 values which are used to modulate the 64 channels of convolutional activations (total of $1*128+128*64=8320$ parameters if $\boldsymbol{\lambda}$ is a scalar). 

\underline{Datasets.} We experiment on both toy datasets derived from CIFAR10/100 and more realistic datasets derived from Kaggle competitions. \textbf{CIFAR10 cat/dog} consists of the cat and dog classes from the CIFAR10 dataset. We also show results for all pairs of CIFAR10 classes in Section~\ref{sec:exp_cifar_pairs}, but find the cat/dog pair is particularly well-suited for additional experiments since it is a challenging classification problem. \textbf{CIFAR100 household electronics vs. furniture} was proposed as a binary dataset with imbalance by \cite{wangImbalanced}. Each class contains 5 classes from CIFAR100: electronics contains clock, computer keyboard, lamp, telephone and television, and furniture contains bed, chair, couch, table and wardrobe. For all CIFAR experiments, we split the data according to their train and test splits. \textbf{Kaggle Dogs vs. Cats} contains 25,000 images of dogs and cats from the Petfinder website \cite{dogs-vs-cats}. We split the data so that each class has 1,000 validation samples (a 92/8 train/validation split). Finally, \textbf{SIIM-ISIC Melanoma} is a classification dataset, which was designed by the Society for Imaging Informatics in Medicine (SIIM) and the International Skin Imaging Collaboration (ISIC) for a Kaggle competition \cite{siim-isic-melanoma-classification}. This is a binary dataset where 8.8\% of the samples belong to the positive (melanoma) class (\ie, $\beta=12$). We follow the procedure of \cite{Fang2023-vq} and combine the 33,126 and 25,331 images from 2020 and 2019 respectively and split the data into an 80/20 train/validation split. \cref{tab:num_samples} outlines the sizes of these. We subsample the minority class to obtain various imbalance ratios $\beta$. Specifically, we test the CIFAR datasets at $\beta=10, 50, 100, 200$, the Kaggle datasets at $\beta=100, 200$.

\underline{Model architectures and training procedure.} We use two model architectures and training procedures, following the literature for each type of dataset. For CIFAR10/100 data, we follow \cite{kini2021labelimbalanced} and use a ResNet-32 model architecture. We find that LCT takes longer to train, so we train for 500 epochs instead of 200 (note: training the baseline models longer does not significantly alter performance and we do this for consistency). We train these models from scratch using an initial learning rate of 0.1 and decreasing this to $1e^3$ and $1e^5$ at 400 and 450 epochs respectively. For the Kaggle datasets, we follow \cite{shwartz-ziv2023simplifying} and use ResNext50-32x4d with weights pre-trained on ImageNet. We train the Melanoma dataset for 10 epochs, following the learning rates and weight decays of \citet{Fang2023-vq} and train the Kaggle Dogs vs. cats models for 30 epochs, following the learning rates and weight decays of other finetuning tasks in \cite{Fang2023-vq}. For all experiments we use a batch size of 128 and gradient clipping with maximum norm equal to 0.5. Additionally, with the exception of Figure~\ref{fig:sam}, we use Stochastic Gradient Descent (SGD) optimization with momentum=0.9. We run all experiments on A5000 GPUs.



\begin{figure*}[tb]
    \centering
    \includegraphics[width=\linewidth]{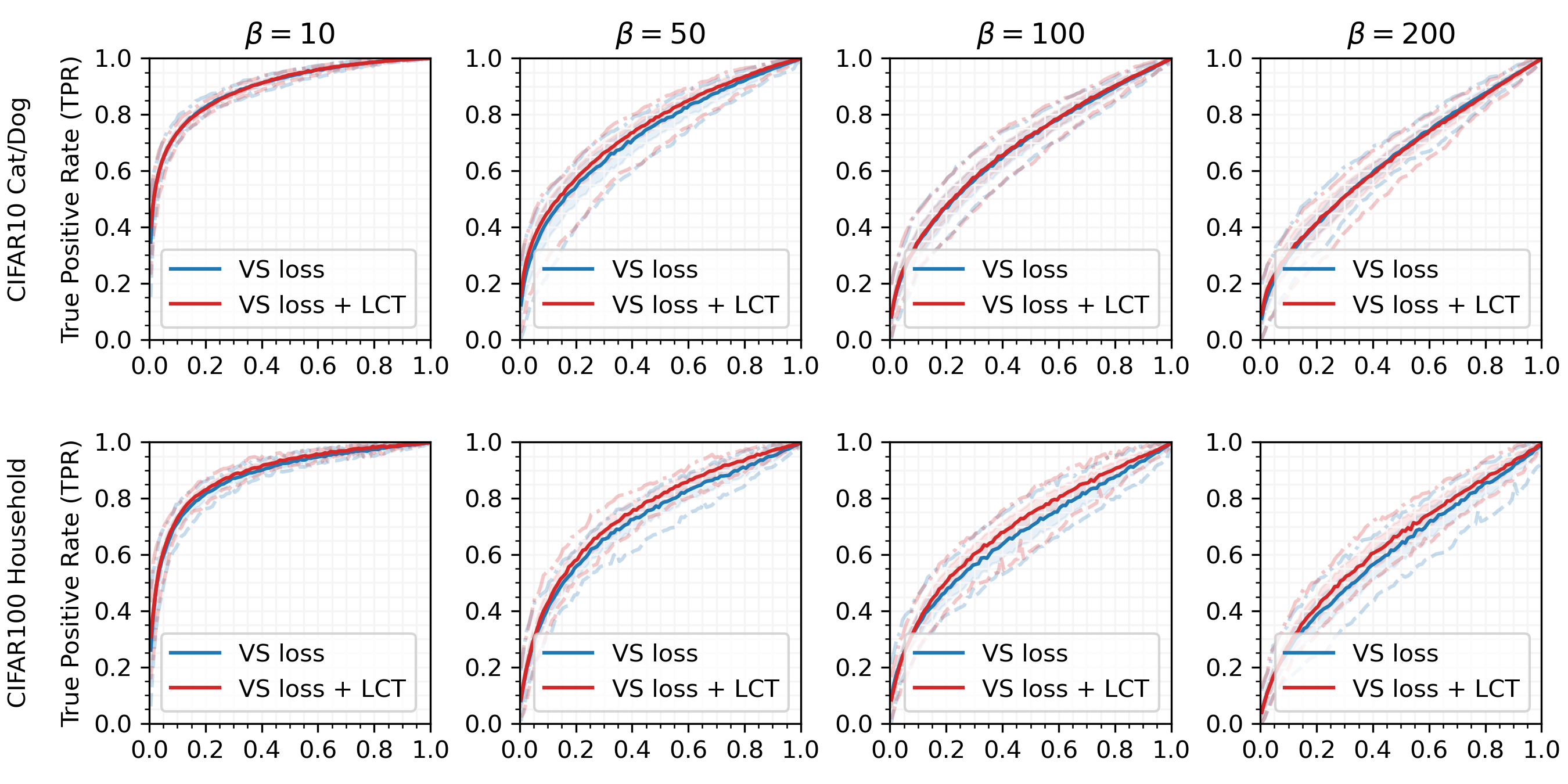}
    \caption{Distribution of ROC curves of models trained with (red) and without (blue) LCT on CIFAR datasets at four different imbalance ratios $\beta$. Solid, dashed, and dotted-dashed curves are mean, minimum, and maximum of the ROC curves respectively. Shaded region is one standard deviation away from the mean. Datasets are tested with $\beta=10, 50, 100, 200$. \textbf{At high imbalance ratios, LCT consistently improves the mean, max, and min of the ROC curves.}}
    \label{fig:cifar_rocs_agg}
\end{figure*}

\begin{figure}[tb]
    \centering
    \includegraphics[width=0.5\linewidth]{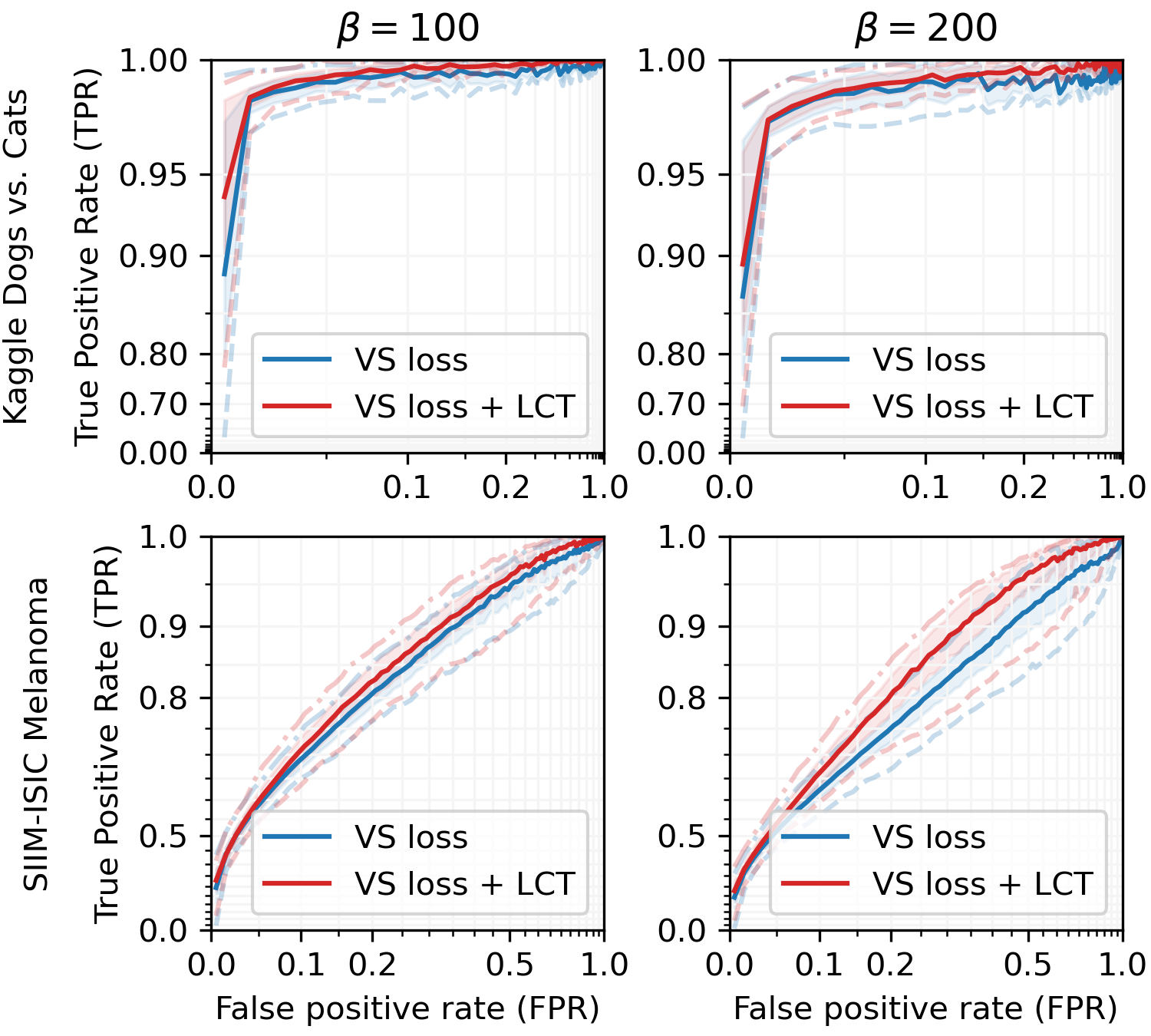}
    \caption{Distribution of ROC curves of models trained with (red) and without (blue) LCT on Kaggle Dogs vs. Cats and SIIM-ISIC Melanoma datasets at two different imbalance ratios $\beta$. Solid, dashed, and dotted-dashed curves are mean, minimum, and maximum of the ROC curves respectively. Shaded region is one standard deviation away from the mean. Results are shown on log scale because all TPR values are compressed towards 1. \textbf{LCT significantly improves performance on Melanoma, especially at high TPRs.}}
    \label{fig:kaggle_rocs_agg}
\end{figure}

\subsection{VS Loss with and without LCT}
\cref{fig:cifar_rocs_agg,fig:kaggle_rocs_agg} compare the performance of the VS loss and VS loss + LCT methods on all four datasets, each at multiple imbalance ratios. For each method, we find the mean, min, max, and standard deviation of the 48 models trained with that method. We calculate these values across false positive rates (FPRs). We also show the distribution of AUCs on the SSIM-ISIC Melanoma dataset in \cref{fig:melanoma_boxplots}. 

\cref{fig:cifar_rocs_agg} shows that at moderate imbalance ratios ($\beta=10$), there is very little variation in the ROC curves obtained by models trained with different hyperparameters. Here the two methods are essentially tied in their performance. However, at high imbalance ratios $\beta=100, 200$ (\cref{fig:cifar_rocs_agg,fig:kaggle_rocs_agg}), the variance in ROC curves becomes much more drastic. Here LCT consistently outperforms the Baseline method. Specifically, LCT improves or ties the max and mean of the ROC curves in all cases, with a large gap in improvement for Melanoma at $\beta=100, 200$.



\subsection{Evaluation over all CIFAR10 pairs}
\label{sec:exp_cifar_pairs}

For a more comprehensive evaluation, we compare the results of VS loss with and without LCT on all 45 pairs of classes chosen from the 10 CIFAR10 classes.
We train each method on each of these binary datasets using 48 different hyperparameter values. We then analyze the aggregated performance of these models by AUC. For each binary dataset, we find the mean, maximum, minimum, and  standard deviation of the AUCs over the hyperparameter values. We then compare Baseline (VS loss) and LCT (VS loss + LCT) for each of these metrics. For example, let $a^{max}_{k, \text{Baseline}}$ and $a^{max}_{k, \text{LCT}}$ be the maximum AUCs achieved on dataset $k$ by LCT and Baseline respectively. Then we obtain 45 of these values for each method (one for each dataset $k$). We then analyze these using a paired t-test and report results in Table~\ref{tab:cifar10}. The table shows that LCT improves performance, increasing the min and mean AUC in nearly all cases, as well as the max AUC in most cases. It also reduces the standard deviation of AUC values. \textbf{This shows that LCT significantly outperforms Baseline on a comprehensive set of datasets.}

\begin{table}[tb]
    \centering
    \small
    \begin{tabular}{|c|cccc|}
    \hline
     & \# LCT $>$ & \# Base. $>$ & Avg. diff & P-value \\
     \hline
        Max & 37 & 8 & 0.004 & 9.1e-03 \\
        Mean & 45 & 0 & 0.010 & 3.9e-17 \\
        Min & 43 & 2 & 0.027 & 2.6e-13 \\
        Std. dev. & 2 & 43 & -0.005 & 3.2e-12 \\
    \hline
    \end{tabular}
    \caption{Differences between AUCs obtained via Baseline (VS loss) and LCT (VS loss + LCT) over 45 binary subsets of CIFAR10. First and second columns show the number of datasets where $a^{max}_{k, \text{LCT}} > a^{max}_{k, \text{Baseline}}$ and vice versa. We also show the average differences in the maximum AUC values over all datasets (\ie, $\frac{1}{|K|} \sum_{k \in K} (a^{\text{max}}_{k, \text{LCT}} - a^{\text{max}}_{k, \text{Baseline}})$) and the P-value for the paired t-test.} 
    \label{tab:cifar10}
\end{table}

\begin{figure}[htbp]
    \begin{minipage}[t]{0.47\textwidth}
        \centering
        \includegraphics[width=\textwidth]{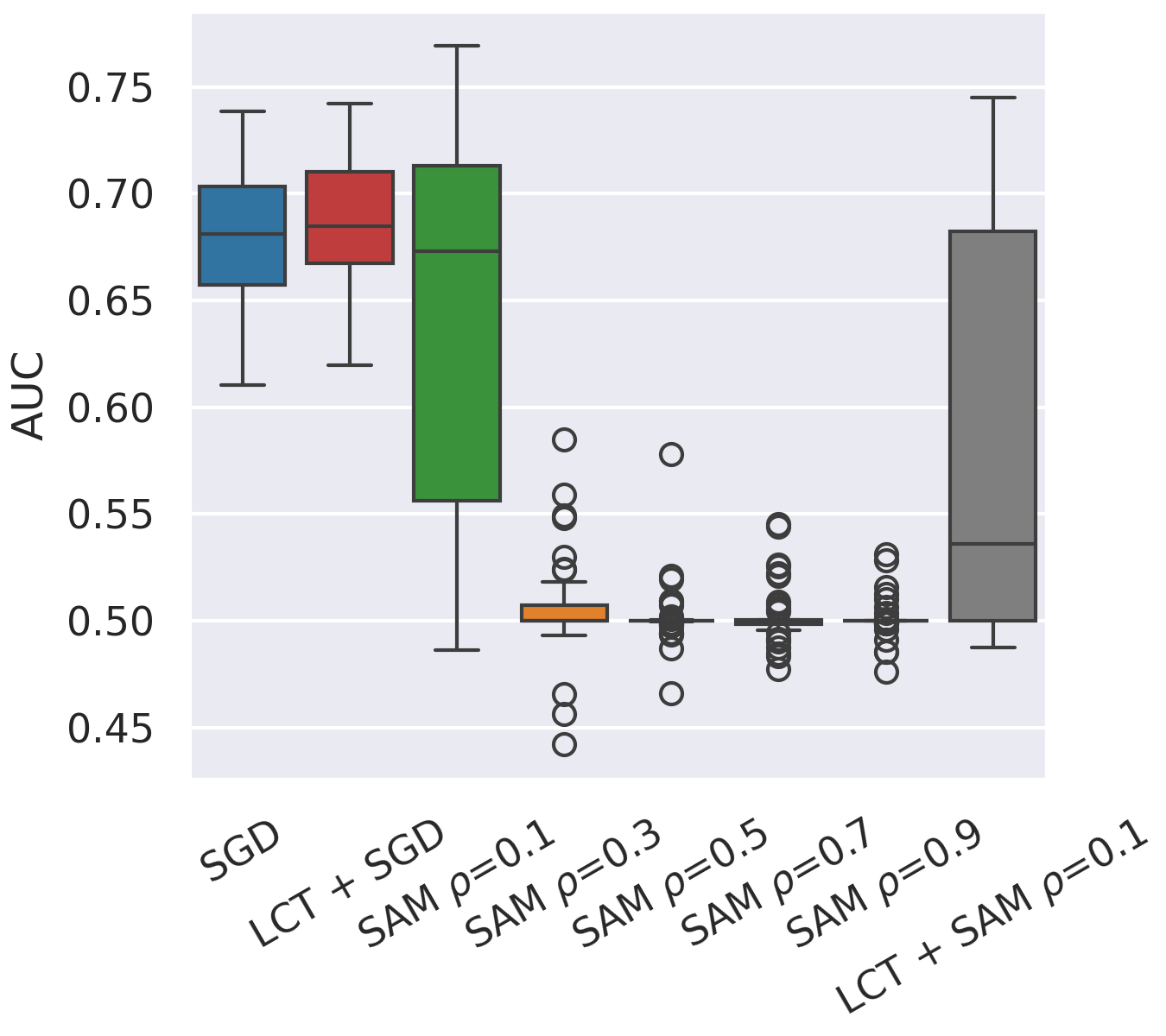}
        \caption{Effects of using SAM optimization. Each boxplot contains results for 48 models trained with different hyperparameter values. Models were trained on CIFAR10 cat/dog $\beta=100$.}
        \label{fig:sam}
    \end{minipage}
    \hfill
    \begin{minipage}[t]{0.47\textwidth}
        \centering
        \includegraphics[width=\textwidth]{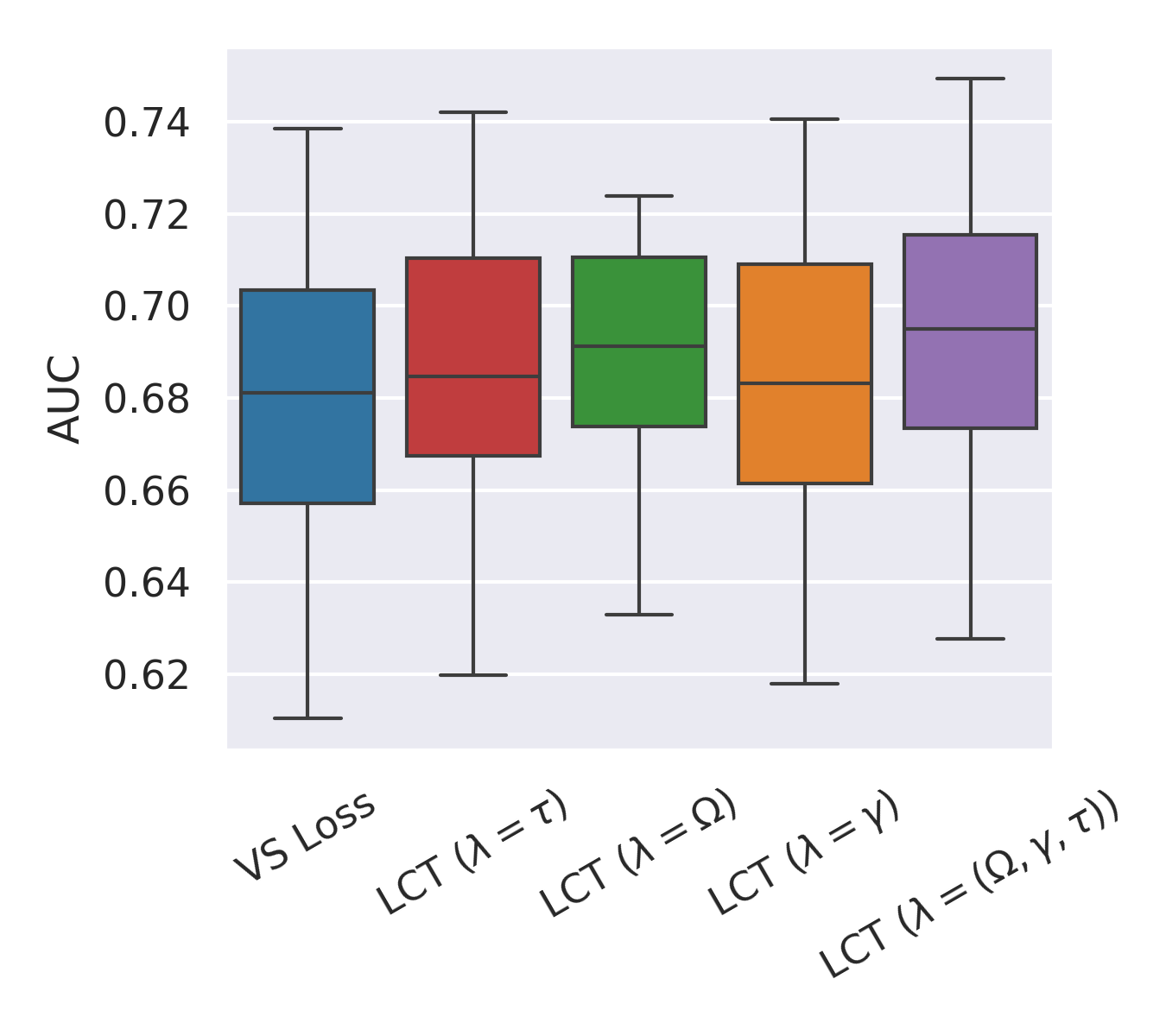}
        \caption{LCT with different $\boldsymbol{\lambda}$. Each boxplot contains results for 27 models with different hyperparameter values. All models were trained on CIFAR10 cat/dog with $\beta=100$.}
        \label{fig:choice_of_lambda}
    \end{minipage}
\end{figure}

\subsection{LCT vs. SAM}
Previous work has shown that Sharpness Aware Minimization (SAM) is effective at optimizing imbalanced datasets in the multi-class setting \cite{foret2021sharpnessaware, rangwani2022escaping}. SAM searches for parameters that lie in neighborhoods of uniformly low loss and usees a hyperparameter $\rho$ to define the size of these neighborhoods. \citet{rangwani2022escaping} shows that SAM can effectively escape saddle points in imbalanced settings by using large values for $\rho$ (\ie, $\rho=0.5, 0.8$). We, however, find that using SAM in this way does not translate to improved performance on a binary dataset. In \cref{fig:sam}, we compare the performance of the models across several hyperparameter values $\rho$. We find that, unlike the multi-class case, in the binary case the best $\rho$ is quite small (\ie, $\rho=0.1$). While training with SAM and $\rho=0.1$ can produce some very strong results, this also exacerbates the amount of variance in the performance. LCT with SGD gives models which are better on average and require less tuning. Note that larger $\rho$ values are absolutely detrimental to training in this setting and lead to models which are effectively naive classifiers. We also tried combining LCT and SAM with $\rho=0.1$ and found this to be less effective than training with LCT and SGD.

\subsection{Choice of $\lambda$}
In this section, we compare different options for which hyperparameters in VS loss to use as $\boldsymbol{\lambda}$ in LCT. We consider setting $\boldsymbol{\lambda}$ to each of the three hyperparameters (using constant values for the other two). 
We also consider setting $\boldsymbol{\lambda}=(\Omega, \gamma, \tau)$. Figure~\ref{fig:choice_of_lambda} compares these options and shows that $\boldsymbol{\lambda}=\tau$ achieves a good compromise between improving the AUC of the models and reducing the variance of results. Furthermore, this choice of $\lambda$ follows our insights from Section~\ref{sec:hyper}.


\section{Conclusion}
In conclusion, we analyze the theoretically well-studied Vector Scaling (VS) loss, and identify that, in practice, its performance exhibits significant variability for binary class imbalance problems with severe imbalance. We mitigate this problem by training over a family of loss functions. We find that this consistently both improves the ROC curves and reduces the model's sensitivity to hyperparameter choices. This improvement comes from the fact that training over a family of loss functions is a proxy for optimizing along different TPR and FPR tradeoffs. Areas of future work include studying how to adapt this method to work on multi-class classification problems under imbalance and regression tasks.



\newpage

\bibliography{references}
\bibliographystyle{iclr2024_conference}

\appendix
\newpage

\section{Binary classification metrics}
\label{sec:appendix_metrics}

In the binary case, we assume that the minority class is the positive class (\ie, the class with label a label of one). For a given classifier, we can categorize the samples in terms of their actual labels and the classifier's predictions as shown in Table~\ref{tab:binary_table}. The remainder of this section defines several metrics used for binary classification.

\begin{table}[H]
    \centering
    \begin{tabular}{|c|c|c|}
        \hline
        & \textbf{Predicted Positive} & \textbf{Predicted Negative} \\
        \hline
        \textbf{Actually Positive} &  \cellcolor{green!20} True Positives (TP) &   \cellcolor{red!20} False Negatives (FN) \\
        \hline
        \textbf{Actually Negative} &  \cellcolor{orange!20} False Positives (FP) & \cellcolor{blue!20} True Negatives (TN) \\
        \hline
    \end{tabular}
    \caption{Categorization of samples in the binary case based on their actual labels (rows) and predicted labels (columns).}
    \label{tab:binary_table}
\end{table}

\subsection{True Positive Rate (TPR) = Minority-class accuracy = Recall}
The True Positive Rate (TPR) is defined as the proportion of actual positive samples which are predicted to be positive. Note that in the binary case, this is equivalent to both the minority-class accuracy and the recall.

\begin{align}
    \text{TPR} = \text{Minority-class acc.} = \text{Recall} =\frac{TP}{TP+FN}
\end{align}

\begin{table}[H]
    \centering
    \begin{tabular}{|c|c|c|}
        \hline
        & \textbf{Predicted Positive} & \textbf{Predicted Negative} \\
        \hline
        \textbf{Actually Positive} & \cellcolor{blue!40} True Positives (TP) & \cellcolor{blue!20} False Negatives (FN) \\
        \hline
        \textbf{Actually Negative} &  False Positives (FP) & True Negatives (TN) \\
        \hline
    \end{tabular}
    \caption{Visualization of TPR calculation. All shaded cells are included in calculation of metric. The numerator is highlighted in dark blue.}
    \label{tab:tpr_table}
\end{table}

\subsection{False Positive Rate (FPR) = 1 - Majority-class accuracy = 1 - TNR}
The False Positive Rate (FPR) is defined as the proportion of actual negative samples which are predicted to be positive. Note that in the binary case, this is equivalent to 1 - the majority-class and 1 - the True Negative Rate (TNR).

\begin{align}
    \text{FPR} = 1 - \text{Majority-class acc.} = \frac{FP}{TN+FP}
\end{align}

\begin{table}[H]
    \centering
    \begin{tabular}{|c|c|c|}
        \hline
        & \textbf{Predicted Positive} & \textbf{Predicted Negative} \\
        \hline
        \textbf{Actually Positive} & True Positives (TP) & False Negatives (FN) \\
        \hline
        \textbf{Actually Negative} &  \cellcolor{blue!40} False Positives (FP) & \cellcolor{blue!20} True Negatives (TN) \\
        \hline
    \end{tabular}
    \caption{Visualization of FPR metric. All shaded cells are included in calculation of metric. The numerator is highlighted in dark blue.}
    \label{tab:fpr_table}
\end{table}

\subsection{Precision}
The precision is defined as the proportion of predicted positive samples which are actually positive. 

\begin{align}
    \text{Precision} =\frac{TP}{TP+FP}
\end{align}

\begin{table}[H]
    \centering
    \begin{tabular}{|c|c|c|}
        \hline
        & \textbf{Predicted Positive} & \textbf{Predicted Negative} \\
        \hline
        \textbf{Actually Positive} & \cellcolor{blue!40} True Positives (TP) & False Negatives (FN) \\
        \hline
        \textbf{Actually Negative} &  \cellcolor{blue!20} False Positives (FP) &  True Negatives (TN) \\
        \hline
    \end{tabular}
    \caption{Visualization of precision metric. All shaded cells are included in calculation of metric. The numerator is highlighted in dark blue.}
    \label{tab:precision_table}
\end{table}

\subsection{Overall Accuracy}
Perhaps the simplest metric is the overall accuracy of the classifier. This is simply the proportion of samples which are correctly classified (regardless of their class). If the test set is imbalanced, a trivial classifier which predicts all samples as negative will achieve a high overall accuracy. Specifically, the overall accuracy of this classifier will be the proportion of negative samples or $\frac{\beta}{1+\beta}$. In class imbalance literature, the overall accuracy is often reported on a balanced test set. In this case, the accuracy is an average accuracy on the positive and negative classes.

\begin{align}
    \text{Overall accuracy} = \frac{TP + TN}{TP + FN + FP + TN}
\end{align}

\begin{table}[H]
    \centering
    \begin{tabular}{|c|c|c|}
        \hline
        & \textbf{Predicted Positive} & \textbf{Predicted Negative} \\
        \hline
        \textbf{Actually Positive} & \cellcolor{blue!40} True Positives (TP) & \cellcolor{blue!20} False Negatives (FN) \\
        \hline
        \textbf{Actually Negative} &  \cellcolor{blue!20} False Positives (FP) &  \cellcolor{blue!40} True Negatives (TN) \\
        \hline
    \end{tabular}
    \caption{Visualization of overall accuracy metric. All shaded cells are included in calculation of metric. The numerator is highlighted in dark blue.}
    \label{tab:overall_acc_table}
\end{table}

\subsection{$F_1$ and $F_\beta$}
In some problems, such as information retrieval, there is only one class of interest (the positive class) and the true negatives can vastly outnumber the other three categories. In this case, a method's effectiveness is determined by 1) how many positive samples it correctly predicted as positive (\ie, the recall) and 2) how many samples are actually positive out of all the samples it predicted as positive (\ie, the precision). The $F_1$ metric measures how well a method can achieve both of these goals simultaneously. Specifically, the $F_1$ measure is the harmonic mean between the precision and recall and is defined as
\begin{align}
    F_1 &= \frac{2 \cdot \text{precision} \cdot \text{recall}}{\text{precision} + \text{recall}}.
\end{align}

The $F_1$ measure assumes that the precision and recall have equal weights; however, sometimes problems have different costs for recall and precision. These asymmetric costs can be addressed by the more general $F_\beta$ metric. Let $\beta$ be the ratio of importance between recall and precision, then $F_\beta$ is defined as \footnote{Note that this $\beta$ differs from the $\beta$ which we defined in the main body.},
\begin{align}
        F_\beta &= \frac{ (1 + \beta^2)  \cdot \text{precision} \cdot \text{recall}}{\beta^2 \cdot \text{precision} + \text{recall}}.
\end{align}

\begin{table}[H]
    \centering
    \begin{tabular}{|c|c|c|}
        \hline
        & \textbf{Predicted Positive} & \textbf{Predicted Negative} \\
        \hline
        \textbf{Actually Positive} & \cellcolor{blue!40} True Positives (TP) & \cellcolor{blue!20} False Negatives (FN) \\
        \hline
        \textbf{Actually Negative} &  \cellcolor{blue!20} False Positives (FP) &  True Negatives (TN) \\
        \hline
    \end{tabular}
    \caption{Visualization of $F_\beta$ metric. All shaded cells are included in calculation of metric. The TP cell is highlighted in dark blue because it is included in both the precision and recall calculation.}
    \label{tab:fbeta_table}
\end{table}

\subsection{G-mean}
The Geometric mean (G-mean or GM) is the geometric mean of the TPR (\ie, sensitivity) and TNR (\ie, specificity) and is defined as follows,
\begin{align}
    GM &= \sqrt{\text{SE} * \text{SP}} \\
    &= \sqrt{\frac{\text{TP}}{\text{TP} + \text{FN}} * \frac{\text{TN}}{\text{TN} + \text{FP}}}.
\end{align}

\begin{table}[H]
    \centering
    \begin{tabular}{|c|c|c|}
        \hline
        & \textbf{Predicted Positive} & \textbf{Predicted Negative} \\
        \hline
        \textbf{Actually Positive} & \cellcolor{blue!40} True Positives (TP) & \cellcolor{blue!20} False Negatives (FN) \\
        \hline
        \textbf{Actually Negative} &  \cellcolor{blue!20} False Positives (FP) &  \cellcolor{blue!40} True Negatives (TN) \\
        \hline
    \end{tabular}
    \caption{Visualization of G-mean metric. All shaded cells are included in calculation of metric. Values in the numerator are highlighted in dark blue.}
    \label{tab:gmean_table}
\end{table}

\subsection{ROC curves and AUROC}
Of course, the number of true positives and true negatives are a trade-off and any method can be modified to give a different combination of these metrics. Specifically, the decision threshold can be modified to give any particular recall. Receiver Operating Characteristic (ROC) curves take this in consideration and show the trade-off of true positive rates and false positive rates over all possible decision thresholds. The area under the ROC curve (AUROC) is calculated using the trapezoid rule for integration over a sample of values and is a commonly used metric \cite{Buda2018-tl}.

\subsection{Precision-Recall Curves and AUPRC}
Similarly, the precision-recall curves show the tradeoff of the recall (on the x-axis) and the precision (on the y-axis) over all possible thresholds. The area under the Precision-Recall Curve (AUPRC) is also calculated using the trapezoid rule.

\section{Number of samples }

\setlength{\tabcolsep}{4.5pt}
\begin{table}[H]
    \centering
    \small
    \begin{tabular}{|c|c|c|c|c|c|c|}
        \hline
        & \multicolumn{4}{|c|}{Train set} & \multicolumn{2}{|c|}{Test set} \\
        \hline
        \multirow{2}{*}{Dataset} & \multirow{2}{*}{\# maj.} &  \# min. & \# min. & \# min. & \# maj. & \# min. \\
        & & ($\beta=10$) & ($\beta=100$) & ($\beta=200$) & & \\
        \hline
        CIFAR10 pair & 5,000 & 500 & 50 & 25 & 1,000 & 1,000\\
        Household & 2,500 & 250 & 25 & 13 & 500 & 500\\
        Dogs vs. Cats & 11,500 & 1,500 & 150 & 75 & 1,000 & 1,000\\
        SIIM-ISIC Melanoma & 41,051 & 4,071* & 410 & 205 & 12,300 & 1,035 \\
        \hline
    \end{tabular}
    \caption{Number of samples in majority class and minority class of the datasets at different imbalance ratios $\beta$. $\beta$ only affects the number of minority-class samples during training. *SIIM-ISIC Melanoma has an imbalance ratio of 12 by default, so we study $\beta=12$ instead of $\beta=10$ in that case.}
    \label{tab:num_samples}
\end{table}
\setlength{\tabcolsep}{6pt}

\section{Effect of evaluation $\lambda$}
\label{sec:eval_lambda}

\begin{figure}[H]
    \centering
    \includegraphics[width=0.5\linewidth]{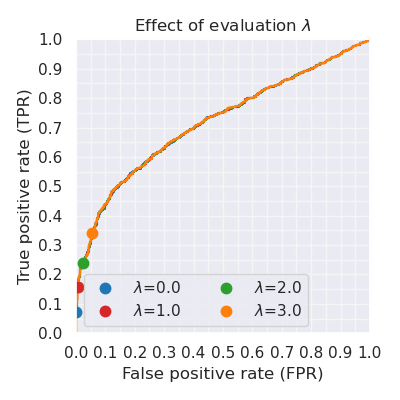}
    \caption{Effect of $\boldsymbol{\lambda}$ value at inference time. Results are shown for one LCT model that has $\boldsymbol{\lambda}=\tau$. Model was trained using the protocol described in \cref{sec: experiments} with $\boldsymbol{\lambda}=\tau$ drawn from a linear distribution on $[0,3]$ with $h_b=0$. Model is evaluated at four $\boldsymbol{\lambda}=\tau$ values: $0.0, 1.0, 2.0$, and $3.0$. For each evaluation $\boldsymbol{\lambda}$, we plot a) the model's TPR/FPR when $t=0.5$ (circle) and b) the ROC curve of the model (curve). \textbf{The evaluation $\boldsymbol{\lambda}$ affects the TPR/FPR for a constant $t$ (\eg, $t=0.5$); however, the ROC curves of models evaluated on different  $\boldsymbol{\lambda}$s are almost identical.} }
    \label{fig:enter-label}
\end{figure}

\section{Linear probability density function}
\label{sec:appendix_linear}
\begin{figure}[H]
    \centering
    \includegraphics[width=\linewidth]{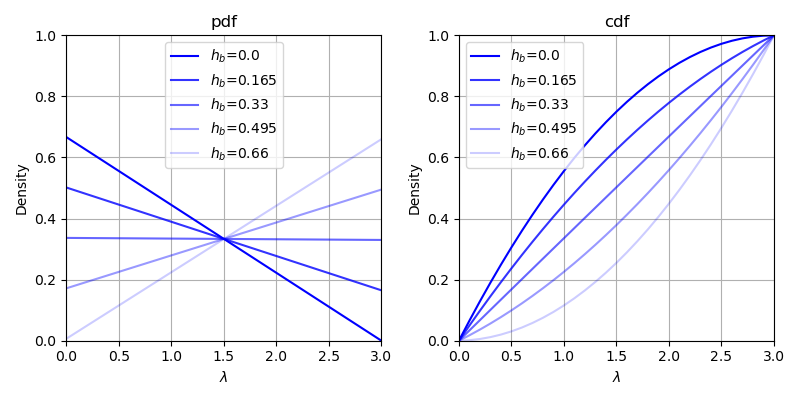}
    \caption{Example PDFs and CDFs of linear distribution with different $h_b$ when $[a,b]=[0, 3]$.}
    \label{fig:linear_dist}
\end{figure}

We use a linear probability distribution to sample $\lambda$ from an interval $[a,b]$. Unlike the triangular distribution, the probability distribution function (PDF) of this distribution can be nonzero at both a and b. \cref{fig:linear_dist} shows several examples of this distribution when $[a,b]=[0, 3.0]$.

To implement this distribution, the user first selects the domain $[a,b]$ and the height of PDF at $b$, $h_b$. The function then calculates $h_a$ so that the area under the PDF equals one. To sample from this distribution, we draw from a uniform(0,1) distribution and use the inverse cumulative distribution function (CDF) of the linear distribution to find a value of $\lambda$. 

Note that this is the uniform distribution on $[a,b]$ when $h_a=h_b$. Additionally, this is a triangular distribution when $h_a=0$ or $h_b=0$.

\section{Simplifying VS Loss}
\label{sec:appendix_simplifying_vs}

Recall the definition of VS loss.
\begin{align}
    \ell_{VS}(y, \mathbf{z}) = -\omega_y \log\left(\frac{e^{\Delta_y z_y + \iota_y}}{\sum_{c\in \mathcal{Y}} e^{\Delta_c z_c + \iota_c}}\right)
\end{align}

Additionally, recall that $\Delta$ and $\iota$ are parameterized by $\gamma$ and $\tau$ as follows.
\begin{align}
    \Delta_c &= \left(\frac{n_c}{n_{max}}\right)^\gamma \\
    \iota_c &= \tau \log \left(\frac{n_c}{\sum_{c' \in C}n_{c'}} \right)
\end{align}

Consider a binary problem with an imbalance ratio $\beta$. Then $\Delta$'s and $\iota$'s are defined as follows
\begin{align}
    \Delta_0 &= 1 & \iota_0=\tau \log \left(\frac{\beta}{\beta+1} \right) \\
    \Delta_1 &= \frac{1}{\beta^\gamma} & \iota_1=\tau \log \left(\frac{1}{\beta+1} \right) \\
\end{align}

We can simplify $\ell_{VS}(0, \mathbf{z})$ as follows. First plug in the $\Delta$'s and $\iota$'s:

\begin{align}
    \ell_{VS}(0, \mathbf{z}) &= - (1-\Omega) \log\left(\frac{e^{z_0+\tau \log (\frac{\beta}{\beta+1})}}{e^{z_0+\tau \log(\frac{\beta}{\beta+1})} + e^{\frac{z_1}{\beta^\gamma} +\tau \log(\frac{1}{\beta+1})}}\right)
\end{align}

Then rewrite $\tau\log(\frac{a}{b})$ as $\tau\log(a) - \tau \log(b)$ and cancel out $e^{\tau \log(\beta+1)}$ from the numerator and denominator.
\begin{align}
    \ell_{VS}(0, \mathbf{z}) &= - (1-\Omega) \log\left(\frac{e^{z_0+\tau \log(\beta) - \tau \log(\beta+1)}}{e^{z_0+\tau \log(\beta) - \tau \log(\beta+1)} + e^{\frac{z_1}{\beta^\gamma} +\tau \log(1) - \tau \log(\beta+1)}}\right) \\
    &= - (1-\Omega) \log\left(\frac{\frac{e^{z_0 + \tau \log(\beta)}}{e^{\tau \log(\beta+1)}}}{\frac{e^{z_0 + \tau \log(\beta)}}{e^{\tau \log(\beta+1)}} + \frac{e^{z_1/\beta^\gamma}}{e^{\tau \log(\beta+1)}}}\right) \\
    &= - (1-\Omega) \log\left(\frac{e^{z_0 + \tau \log(\beta)}}{e^{z_0 + \tau \log(\beta)} + e^{z_1/\beta^\gamma}}\right)
\end{align}

Then use the following two facts to a) simplify the term inside the log and b) rewrite the log term: 
\begin{align}
    &\text{a) } \frac{e^a}{e^a+e^b} = \frac{e^a}{e^a+e^b} \frac{\frac{1}{e^a}}{\frac{1}{e^a}} = \frac{1}{1+e^{b-a}} \\
    &\text{b) } \log\left(\frac{1}{1+e^{b-a}}\right) = \log(1) - \log(1+e^{b-a}) = - \log(1+e^{b-a})
\end{align}

\begin{align}
    \ell_{VS}(0, \mathbf{z}) &= (1-\Omega)\log\left(1+ e^{z_1/\beta^\gamma - z_0 - \tau \log(\beta)} \right)
\end{align}

Finally, we move the $\tau\log(\beta)$ term outside of the exponent
\begin{align}
    \ell_{VS}(0, \mathbf{z}) &= (1-\Omega)\log\left(1+ \frac{1}{\beta^\tau} e^{z_1/\beta^\gamma - z_0} \right).
\end{align}

We follow a similar process for $\ell_{VS}(1, \mathbf{z})$ and get

\begin{align}
    \ell_{VS}(1, \mathbf{z}) &= \Omega\log\left(1 + \beta^\tau e^{z_0 - z_1/\beta^\gamma}\right).
\end{align}
\section{Partial derivatives of VS-Loss with respect to $\Omega, \tau, \gamma$}
\label{sec:appendix_hyper_partial}

Then the partial derivatives of the loss when $y=1$ with respect to $\Omega, \gamma$ and $\tau$ are as follows.

\begin{align}
    \frac{\partial}{\partial \Omega} \ell_{VS}(1, \mathbf{z}) = \log\left(1 + \beta^\tau e^{z_0 - z_1/\beta^\gamma}\right) \\
    \frac{\partial}{\partial \gamma} \ell_{VS}(1, \mathbf{z}) = \frac{\Omega z_1 e^{z_0 - z_1/\beta^\gamma}\beta^{\tau - \gamma} \log(\beta) } {1 + \beta^\tau e^{z_0 - z_1/\beta^\gamma}} \\
    \frac{\partial}{\partial \tau} \ell_{VS}(1, \mathbf{z}) = \frac{\Omega e^{z_0-z_1/\beta^\gamma}\beta^\tau \log(\beta)} {1 + \beta^\tau e^{z_0 - z_1/\beta^\gamma}}
\end{align}

In particular,

\begin{equation}
\frac{\partial}{\partial \gamma} \ell_{VS}(1, \mathbf{z}) = \frac{z_1}{\beta^\gamma} \frac{\partial}{\partial \tau} \ell_{VS}(1, \mathbf{z})
\label{eq:gamma_tau_partials}
\end{equation}

and for balanced classes

\[
\frac{\partial}{\partial \gamma} \ell_{VS}(1, \mathbf{z}) = z_1 \frac{\partial}{\partial \tau} \ell_{VS}(1, \mathbf{z}) \;.
\]

\section{Full derivation for break-even points}
\label{sec:appendix_break-even}

With VS loss defined in \cref{eq:vs0,eq:vs1}, denote $\alpha=z_1/\beta^\gamma - (z_0 + \tau \log\beta)$, setting $\ell_{VS}(1,\mathbf{z}) = \ell_{VS}(0,\mathbf{z})$ gives us 
\begin{align}
(1+e^{-\alpha})^\Omega=(1+e^\alpha)^{1-\Omega}.
\end{align}

Consider the function $f(\alpha)=(1+e^{-\alpha})^\Omega-(1+e^\alpha)^{1-\Omega}$. Since $0\leq\Omega\leq1$, it is clear that $f(\alpha)$ is a continuous and monotonically decreasing function. Also, $f(\alpha)\to+\infty$ as $\alpha\to-\infty$, and $f(\alpha)\to-\infty$ as $\alpha\to+\infty$. These imply that $f(\alpha)=0$ has exactly one unique solution, depended on $\Omega$. Let $\alpha_\Omega$ be the solution such that $f(\alpha_\Omega)=0$. 

Notice that when $\Omega=0.5$, the unique solution $\alpha_\Omega$ = 0. We can also see that if $\Omega>0.5$,  we have $f(0)=2^\Omega-2^{1-\Omega}>0$, and since $f(\alpha)$ is monotonically decreasing, we know that the solution $\alpha_\Omega$ should be strictly larger than 0. Otherwise, (\ie, $\Omega<0.5$), we have $\alpha_\Omega < 0$.


Suppose we have found the $\alpha_\Omega$ such that $f(\alpha_\Omega)=0$, this means that the break-even point should have $z_1/\beta^\gamma = z_0 + \tau \log\beta + \alpha_\Omega$. Compared with the regular cross-entropy break-even point, where $z_1=z_0$, we are adding biases here by the terms $\beta^{-\gamma}$, $\tau\log\beta$, and $\alpha_\Omega$, dependent on $\gamma, \tau, \Omega$, respectively. These terms change the margin between $z_0$ and $z_1$, which poses a bias for samples on the break-even point. Specifically, for a break-even point sample (a sample that the trained model is totally confused with), the model still prefers a positive output with $z_1>z_0$ if $\beta^\gamma>1$, $\tau\log\beta>0$, and/or $\alpha_\Omega > 0$, which translate to $\gamma>0$, $\tau>0$, and/or $\Omega>0.5$, respectively.

\subsection{Numerical Example}
\label{sec:appendix_break_even_numerical}

Since VS loss introduces a bias term $\tau\log (\beta)$ to $z_0$ during training, the model will learn to consistently shrink its $z_0$ and/or increase its $z_1$ outputs to match this bias. Specifically, in the case $\Omega=0.5, \gamma=0$, but arbitrary $\tau$, the will re-calibrate to give output $\mathbf{z} = (x, x + \tau \log (\beta))$ where $x$ is arbitrary to samples it is completely unsure about. Consequently, the softmax output $p_1$ associated with the break-even points will be larger. For example, when $\tau=0$ (regular cross-entropy), the softmax $p_1$ associated with the break-even points is 0.5. When $\tau>0$, the softmax score associated with the break-even points is $\frac{\beta^\tau}{1 + \beta^\tau}$ (\eg, 0.99 when $\beta=10$ and $\tau=2$).

\section{Contour plots of effect of $\Omega, \gamma, \tau$ on VS loss}
\label{sec:appendix_hyper_contour}

\begin{figure}[H]
    \centering
    \includegraphics[width=0.8\linewidth]{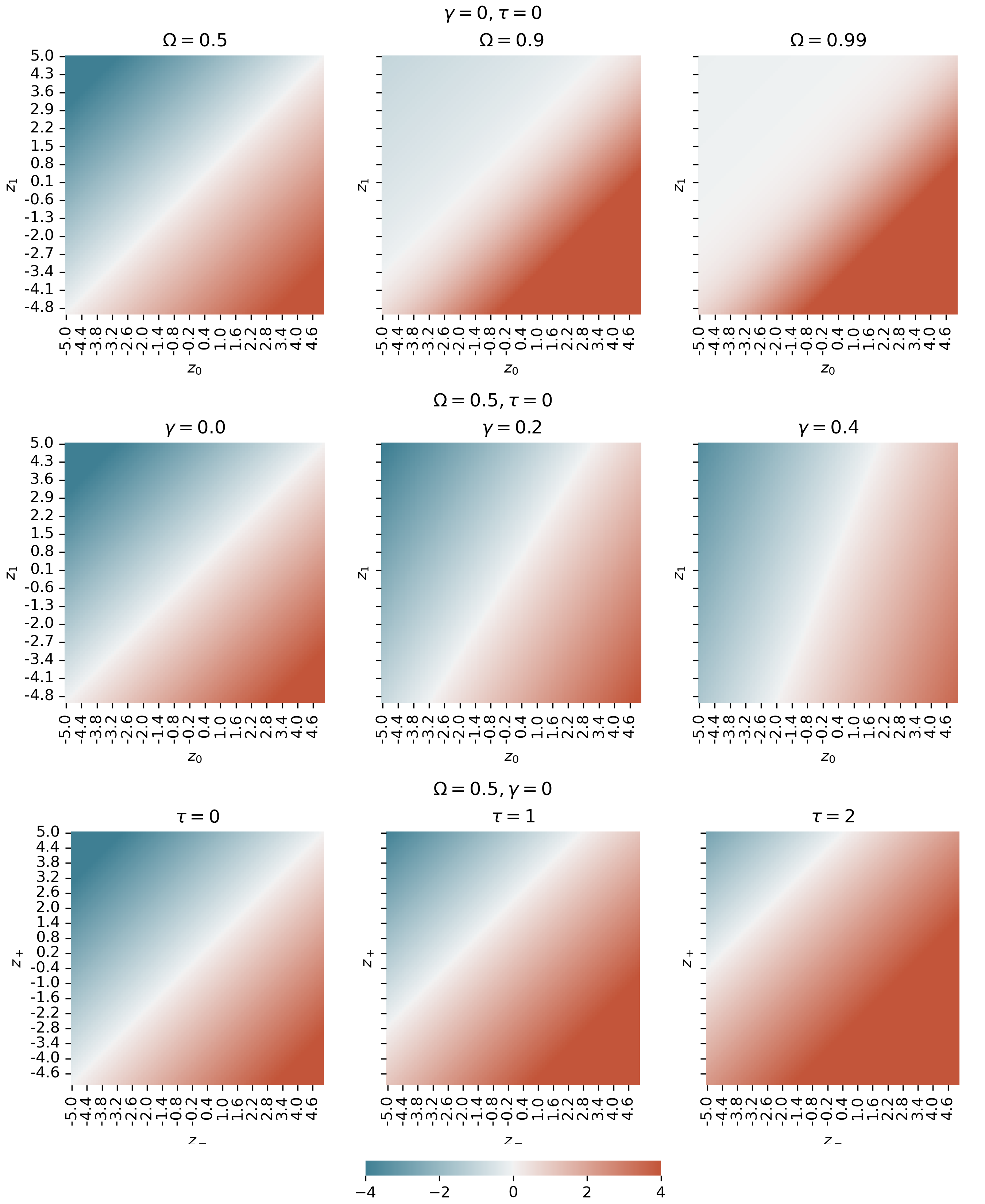}
    \caption{Effect of hyperparameter values on $\ell_{VS}(1,\mathbf{z})- \ell_{VS}(0,\mathbf{z})$ over $z_0, z_1 \in [-5, 5]$ for $\beta=10$. Top, middle, and bottom rows show results for different values of $\Omega$, $\gamma$, and $\tau$ respectively. A value of zero (white) represents the ``break even'' points.}
    \label{fig:break_even}
\end{figure}

In this section we consider the effect of hyperparameter values for $\Omega, \gamma, \tau$ on the values of the loss functions over a set of $\mathbf{z}$ for a binary problem. Specifically, we consider the difference of the loss when the label is 1 and 0 (\ie, $\ell_{VS}(1,\mathbf{z}) - \ell_{VS}(0,\mathbf{z})$). This gives insights into how the model prioritizes different classes over the domain of $\mathbf{z}$. We plot this for several combinations of hyperparameter values in Figure~\ref{fig:break_even}. In particular, we are interested in the set of ``break-even'' points or the points of $\mathbf{z}$ where $\ell_{VS}(1,\mathbf{z}) = \ell_{VS}(0,\mathbf{z})$ because this gives insights into the thresholds which a model with that loss function optimizes over.

\subsection{Regular cross-entropy}
First consider regular cross-entropy loss (\ie, $\Omega=0.5, \gamma=0, \tau=0$). Here $\ell_{VS}(1,\mathbf{z}) = \ell_{VS}(0,\mathbf{z})$ when $z_1=z_0$. This is shown on the top left column of Figure~\ref{fig:break_even}, where the line corresponding to $z_1=z_0$ is white, indicating that the losses are equal along this line. If $z_1>z_0$ (\ie, $\hat{y} = 1$), then the difference of the losses is negative (\ie, $\ell_{VS}(1,\mathbf{z}) < \ell_{VS}(0,\mathbf{z})$). This makes sense since the model should receive a bigger loss for an incorrect prediction. Notice that this plot is symmetric along the line $z_1=z_0$.

\subsection{Varying $\Omega$ (Weighted cross-entropy loss)}
Next consider VS loss with arbitrary weights $\Omega$, but fixed $\gamma=0, \tau=0$. Note that this is equivalent to weighted cross-entropy with class 1 having weight $\Omega$ and class 0 having weight $1-\Omega$. We include contour plots of this case in the top row of Figure~\ref{fig:break_even} and simplify the equation of equal losses below. Unfortunately there is no analytic solution in this case, but we provide more insights about this case in Subsection~\ref{sec:omega_vs_softmax}.

\begin{align}
    \ell_{VS}(1,\mathbf{z}) &= \ell_{VS}(0,\mathbf{z}) \\
    -\Omega \log\left(\frac{e^{z_1}}{e^{z_0} + e^{z_1}}\right) &= -(1-\Omega) \log\left(\frac{e^{z_0}}{e^{z_0} + e^{z_1}}\right) \\
    \left(\frac{e^{z_1}}{e^{z_0}+e^{z_1}}\right)^\Omega &= \left(\frac{e^{z_0}}{e^{z_0}+e^{z_1}}\right)^{1-\Omega} \\
    e^{\Omega z_1} &= e^{(1-\Omega)z_0}\left(e^{z_0}+e^{z_1}\right)^{2\Omega-1} \\
    z_1 &= \left(\frac{1-\Omega}{\Omega}\right)z_0 + \left(\frac{1}{\Omega}\right)\log (e^{z_0}+e^{z_1})^{2\Omega-1}
\end{align}

\subsection{Varying $\gamma$}
Next consider VS loss with with arbitrary values of $\gamma$, but fixed $\tau=0,\Omega=0.5$. Recall from the previous section that for a binary problem with an imbalance ratio $\beta$, $\Delta_0 = 1$ and $\Delta_1=1/\beta^\gamma$. Then

\begin{align}
    \ell_{VS}(1,\mathbf{z}) &= \ell_{VS}(0,\mathbf{z}) \\
    -0.5 \log(\frac{e^{z_1/\beta^\gamma}}{e^{z_0} + e^{z_1/\beta^\gamma}}) &= -0.5 \log(\frac{e^{z_0}}{e^{z_0} + e^{z_1/\beta^\gamma}}) \\
    z_1 &= \beta^\gamma z_0 \\
\end{align}

Thus changing $\gamma$ equates to rotating the line of break-even points as seen on the middle row of Figure~\ref{fig:break_even}. 

\subsection{Varying $\tau$}
Finally consider VS loss with with arbitrary values of $\tau$, but fixed $\gamma=0,\Omega=0.5$. Recall from the previous section that for a binary problem with an imbalance ratio $\beta$, $\iota_0=\tau \log \left(\frac{\beta}{1+\beta} \right)$ and $\iota_1=\tau \log \left(\frac{1}{1+\beta} \right)$. Then
\begin{align}
    \ell_{VS}(1,\mathbf{z}) &= \ell_{VS}(0,\mathbf{z}) \\
    -0.5 \log\left(\frac{e^{z_1 + \tau \log \left(\frac{1}{1+\beta} \right)}}{e^{z_0 + \tau \log \left(\frac{\beta}{1+\beta} \right)} + e^{z_1 + \tau \log \left(\frac{1}{1+\beta} \right)}}\right) &= -0.5 \log\left(\frac{e^{z_0 + \tau \log \left(\frac{\beta}{1+\beta} \right)}}{e^{z_0 + \tau \log \left(\frac{\beta}{1+\beta} \right)} + e^{z_1 + \tau \log \left(\frac{1}{1+\beta} \right)}}\right) \\
    z_1 + \tau \log \left(\frac{1}{1+\beta} \right) &= z_0 + \tau \log \left(\frac{\beta}{1+\beta} \right) \\
    z_1 &= z_0 + \tau\log\left(\frac{\frac{\beta}{1+\beta}}{\frac{1}{1+\beta}} \right) \\
    z_1 &= z_0 + \tau\log\left(\beta \right) \\
\end{align}

Thus changing $\tau$ equates to shifting the line of break-even points as seen on the bottom row of Figure~\ref{fig:break_even}.

\subsection{$\Omega$ as a function of softmax scores}
\label{sec:omega_vs_softmax}

Recall that each model outputs a vector of logits $\mathbf{z}$ and that $\hat{y}=\arg\max \mathbf{z}$. In the binary case, this is equivalent to computing the softmax score for $z_1$ and thresholding this score at 0.5 as shown below.
\begin{align}
    p_1 &= \frac{e^{z_1}}{e^{z_0}+e^{z_1}} \\
    \hat{y} &= 
    \begin{cases}
        1 & \text{if } p_1 \geq 0.5 \\
        0 & \text{otherwise}
    \end{cases}
\end{align}

\begin{figure}[H]
    \centering
    \includegraphics[width=\linewidth]{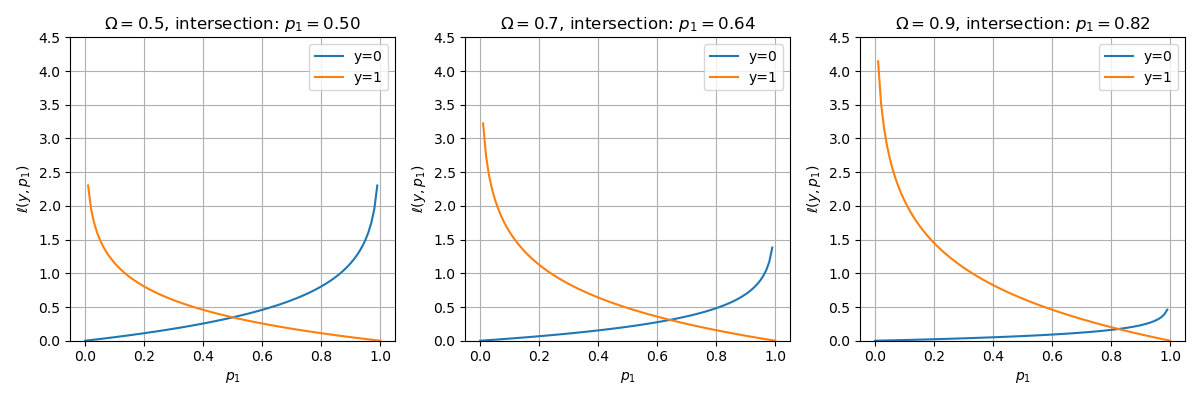}
    \caption{Weighted cross entropy loss over different values of $\Omega$ and softmax scores $p_1$, which correspond to class 1. Note that $p_0=1-p_1$.}
    \label{fig:break_even_omega}
\end{figure}

Because 0-1 loss is not differentiable, models are instead optimized for a proxy function like cross-entropy. In traditional cross-entropy (\ie, $\Omega=0.5$), the loss function is symmetric such that $\ell(0, p_1) = \ell(1, -p_1)$. Figure~\ref{fig:break_even_omega} shows how this threshold varies for different values of $\Omega$. As $\Omega$ increases, the line $y=1$ becomes steeper while $y=0$ becomes less steep and the value of $p_1$ where the losses intersect increases. 
Note that the point of intersection is 
\begin{align}
    p_1^\Omega = (1-p_1)^{1-\Omega}.
\end{align}

Additionally, note that changing $\gamma$ or $\tau$ does not change the point $p_1$ where the losses intersect.

\end{document}